\documentclass[letterpaper]{article} 
\usepackage{aaai2026}  
\usepackage{times}  
\usepackage{helvet}  
\usepackage{courier}  
\usepackage[hyphens]{url}  
\usepackage{graphicx} 
\usepackage{natbib}  
\usepackage{caption} 
\usepackage{algorithm}
\usepackage{algorithmic}

\usepackage{amsmath}
\usepackage{amssymb}
\usepackage{amsfonts}

\usepackage{booktabs}
\usepackage{multirow}
\usepackage{array}

\usepackage{xcolor}

\usepackage{newfloat}
\usepackage{listings}
\DeclareCaptionStyle{ruled}{labelfont=normalfont,labelsep=colon,strut=off} 
\floatstyle{ruled}
\newfloat{listing}{tb}{lst}{}
\floatname{listing}{Listing}
\nocopyright
\usepackage{hyperref}

\title{Model Whisper: Steering Vectors Unlock\\ Large Language Models' Potential in Test-time}
\author{
    Xinyue Kang\textsuperscript{\rm 1},
    Diwei Shi\textsuperscript{\rm 2,\rm 1},
    Li Chen\textsuperscript{\rm 1}\thanks{Corresponding author.}
}

\affiliations{
    \textsuperscript{\rm 1}Tsinghua Shenzhen International Graduate School, Tsinghua University\\
    \textsuperscript{\rm 2}Research Institute of Tsinghua University in Shenzhen\\

    kangxy24@mails.tsinghua.edu.cn, sdw18701584492@163.com, lic87766@gmail.com
}

\usepackage{bibentry}

\begin{document}

\maketitle

\begin{abstract}
It is a critical challenge to efficiently unlock the powerful reasoning potential of Large Language Models (LLMs) for specific tasks or new distributions. Existing test-time adaptation methods often require tuning model parameters, which is not only computationally expensive but also risks degrading the model's pre-existing abilities.To address this, we introduce a lightweight component, \textit{Test-Time Steering Vectors} (TTSV), which is prepended to the input while keeping the LLM's parameters entirely frozen. By optimizing the TTSV on test data to minimize the model's output entropy, we steer the model towards an internal state of higher confidence, activating its inherent abilities most relevant to the current task. TTSV is both lightweight and highly efficient to optimize, making it a true plug-and-play enhancement. Extensive experiments validate our approach's effectiveness on both base models and reasoning-enhanced models. For instance, on the MATH500 task, TTSV achieves a 45.88\% relative performance gain on the Qwen2.5-Math-7B model and a 16.22\% relative gain on the Qwen3-4B model. Furthermore, our approach exhibits robust generalization, with its steering vectors proving highly transferable across diverse tasks.
\end{abstract}

\begin{links}
    \link{Code}{https://github.com/kkkkxy/TTSV}
\end{links}

\section{Introduction}

Large Language Models (LLMs) have demonstrated remarkable capabilities \cite{jaech2024openai,guo2025deepseek}, evolving into powerful general-purpose problem solvers. Trained on massive corpora of text and code, these models internalize a wealth of world knowledge and complex reasoning patterns \cite{wang2024knowledge}. However, this immense potential often remains latent. When confronted with a specific task or a novel data distribution, a model may fail to automatically or optimally deploy its intrinsic capabilities, even though they exist \cite{xu2025towards}. This gap between the model's latent potential and its actual performance presents a key challenge to fully realizing its value.

Prevailing methods for guiding large models toward desired behaviors primarily involve post-training, such as fine-tuning the entire model via Reinforcement Learning from Human Feedback (RLHF) \cite{bai2022training} or Reinforcement Learning with Verifiable Rewards (RLVR) \cite{guo2025deepseek}. These approaches, though effective, demand substantial computational resources and extensive, costly human-annotated data. Test-time adaptation (TTA) offers a more lightweight alternative by dynamically adapting the model to unlabeled test data. Yet, despite their proven efficacy, prevailing TTA strategies still fundamentally depend on parameter modification, often involving the update of all model parameters at test time, as seen in approaches like Test Time Reinforcement Learning (TTRL) \cite{zuo2025ttrl} and One-shot Entropy Minimization (EM) \cite{gao2025em}. This dependency on parameter modification gives rise to two critical challenges: 1) the considerable computational expense of gradient calculation and backpropagation, and 2) the inherent risk of catastrophic forgetting, whereby adapting to a new domain might unintentionally erase the model's invaluable, pretrained general abilities \cite{liu2021ttt++}. 

The parameter-efficient TTA method SLOT \cite{hu2025slot} offers an alternative by introducing a learnable parameter vector to the final hidden layer before the output head, acting as a form of output-level calibration. However, due to its limited influence on the final computational step, the entire upstream multi-layer reasoning process remains unguided, restricting its performance.

The question then becomes: how can we guide the LLM's entire reasoning process from its inception, activating its latent potential, while still keeping its core parameters untouched? This requires a paradigm shift from model modification to holistic model activation—unlocking rather than reshaping its inherent capabilities. While Prompt Engineering \cite{sahoo2024systematic} also intervenes at the input stage and preserves model parameters, it essentially ``shouts'' at the model through discrete, human-readable text. The expressive power of discrete vocabulary consequently limits its effectiveness, whereas the model's native representation space is continuous and capable of encoding denser, more fine-grained information.

Accordingly, we propose Model Whispering. Instead of ``shouting'', we learn to directly ``whisper'' a cryptic instruction into the model's continuous embedding space. This signal, though opaque to humans, is information-rich and designed to precisely guide the model's internal state into the optimal configuration for solving the current task.

We realize the ``Model Whispering" concept as a lightweight component named the Test-Time Steering Vectors (TTSV). The TTSV is a set of short, optimizable vectors that are directly prepended to the front of the input text's embedding sequence. This placement is deliberate. By intervening at the very beginning, TTSV guides the entire multi-layer reasoning process, a fundamental distinction from output-level adjustments like SLOT. Specifically, as the prepended vectors participate in the computation within the attention modules across all layers, the output of each module is effectively a linear combination of its original output and a learned bias vector \cite{petrov2023prompting}. This bias vector serves to guide the model's computational trajectory into an optimal configuration, all while preserving the relative attention patterns among the original input tokens.

To optimize TTSV, our approach is guided by a core premise: a model that successfully activates its relevant capabilities will exhibit greater certainty in its predictions. Since output entropy directly measures this predictive uncertainty, we define the optimization goal for the TTSV as minimizing the output entropy. This objective compels the model to converge towards a more definitive and confident judgment, in turn achieving the precise activation of its latent potential. The entire framework is remarkably efficient and non-invasive, requiring just a few gradient updates to the minuscule TTSV and rendering it a genuine plug-and-play enhancement.

We conduct experiments on a wide range of LLMs and benchmarks to validate the effectiveness of our approach. For instance, on the MATH500 dataset, TTSV achieves a 45.88\% relative performance gain on the Qwen2.5-Math-7B model (lifting its score from 51.00\% to 74.40\%), and a 16.22\% relative gain on the Qwen3-4B reasoning model (lifting its score from 51.80\% to 60.20\%). Furthermore, the approach demonstrates robust generalization capabilities. The TTSV optimized on MATH500, for example, boosts performance on the unseen AMC23 dataset by a substantial 20.00\%. This  interesting phenomenon highlights a key property of our method: a TTSV trained on one task can be effectively transferred to another, consistently unlocking significant performance improvements without re-optimization.

Our contributions can be summarized as follows.

\begin{itemize}
    \item We propose a novel, light-weight paradigm for TTA. By optimizing an external, plug-and-play component (TTSV) prepended to the input, our method steers the model's entire reasoning process from its inception. This characteristic fundamentally distinguishes our approach from methods that either modify model weights or only perform output-level calibration, offering a new, non-invasive, and more holistic path for guiding LLMs.
    \item We demonstrate that TTSV can be effectively optimized using a simple, label-free objective: entropy minimization. This process successfully activates the model's latent abilities without needing complex reward functions or access to internal gradients.
    \item Extensive experiments demonstrate our method's high effectiveness, strong generalization, and preservation of inherent capabilities. 
\end{itemize}

\begin{figure*}[t]
\centering
\includegraphics[width=1.0\textwidth]{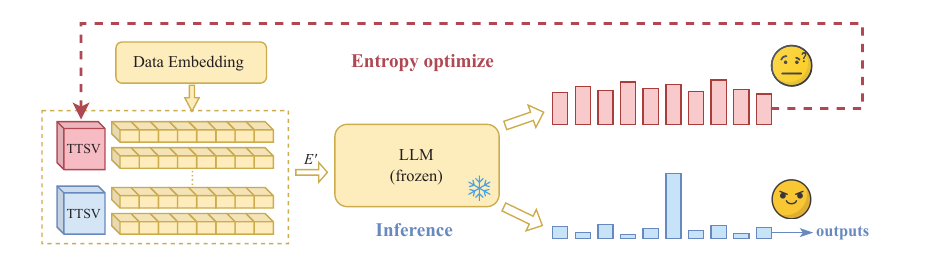} 
\caption{The overall framework of our proposed Model Whisper method. 
The \textbf{upper path (red)} illustrates the optimization phase, where the TTSV is iteratively updated by minimizing the model's output entropy. 
The \textbf{lower path (blue)} shows the inference phase, where the fixed, optimized TTSV guides the model to make low-entropy, high-confidence predictions.}
\label{framework}
\end{figure*}

\section{Related Work}

\subsection{Test-Time Adaptation (TTA)}
TTA aims to improve a pre-trained model's performance by adapting it to unlabeled test instances, thereby addressing distribution shifts encountered during deployment \cite{liang2025comprehensive}. Pioneered in computer vision, foundational TTA approaches often focused on minimizing prediction uncertainty. A prominent example is TENT \cite{wang2020tent}, which minimizes the entropy of predictions to adapt batch normalization layers. Other methods, like Test-Time Training (TTT) \cite{sun2020test}, employ self-supervised tasks such as rotation prediction for adaptation.

More recently, the principles of TTA have been applied to LLMs. For instance, TTRL \cite{zuo2025ttrl} generates pseudo-labels for test data via a majority vote and then performs full-parameter test-time training using reinforcement learning. Another approach, SLOT \cite{hu2025slot}, optimizes a parameter vector added to the model's final hidden layer by performing next-token prediction on the input prompt itself. While these methods have demonstrated efficacy, they face fundamental challenges. The risk of catastrophic forgetting during parameter modification \cite{liu2021ttt++,zhao2023pitfalls}, and the question of how to more deeply guide the model to adapt to a new distribution rather than merely calibrating its output, remain open problems. Consequently, designing lightweight, efficient, and safe TTA methods specifically tailored for LLMs remains a key research direction.

\subsection{Parameter-Efficient Fine-Tuning (PEFT)}

PEFT methods specialize pretrained models for downstream applications by optimizing a minimal subset of parameters during a dedicated fine-tuning phase, while leaving the vast majority of the original LLM weights frozen. Notable examples of these techniques include Adapter Tuning \cite{houlsby2019parameter}, LoRA \cite{hu2022lora}, Prompt Tuning \cite{lester2021power}, and Prefix-Tuning \cite{li2021prefix}.

A common principle unites these methods: they are employed during a training phase, often in a supervised setting, to embed new and task-specific knowledge into the trainable components. The goal is to effectively teach the model a new skill or specialize its behavior for a particular downstream task. In stark contrast to this training-centric philosophy, our approach has a different objective: to activate and steer the model's pre-existing capabilities at test-time. Operating on unlabeled data, our method focuses on dynamically unlocking what the model already knows, rather than instilling new, task-specific information.

\subsection{Entropy-based Optimization}

The concept of entropy, as a measure of uncertainty, serves as a powerful, dual-purpose objective in machine learning. On one hand, entropy maximization is often employed as a regularization technique, particularly in reinforcement learning (RL), where the expected reward objective is augmented with a term designed to increase the policy's entropy \cite{ziebart2008maximum}. The goal is to promote exploration by encouraging uncertainty in the model's outputs or actions \cite{haarnoja2018soft,eysenbach2021maximum,wang2025reinforcement}. This approach is well-suited for scenarios requiring extensive exploration, such as learning a policy from scratch. However, for a capable LLM that already possesses a strong base of knowledge and reasoning patterns, forcing such randomized exploration can be counterproductive. It may destabilize the model's coherent thought processes, leading to less reliable and structured outputs.

On the other hand, entropy minimization has a long history as a principle for encouraging model confidence and exploiting learned knowledge. It is a well-established regularization technique in semi-supervised learning \cite{grandvalet2004semi, berthelot2019mixmatch, niu2014information}, domain adaptation \cite{roy2019unsupervised, saito2019semi}, and few-shot learning \cite{dhillon2019baseline, kim2022infonerf}. More recently, a number of LLM-related works have also leveraged entropy minimization for optimization \cite{gao2025em,agarwal2025unreasonable,prabhudesai2025maximizing}.

Since TTA is applied to an LLM with substantial latent abilities, the primary challenge is not to discover new strategies but to reliably activate the correct ones. Therefore, we argue that minimizing entropy, thereby reducing uncertainty and steering the model towards a state of high confidence, is a direct and effective strategy for unlocking the task-relevant knowledge within a pre-trained LLM.

\begin{table*}[ht]
\centering
\begin{tabular}{lccccc}
\toprule
\textbf{Model} & \textbf{MATH500} & \textbf{Minerva Math} & \textbf{Olympiad} & \textbf{AMC23} & \textbf{Avg.} \\
\midrule
Qwen2.5-Math-7B    & 51.00 & 12.90 & 16.70 & 42.50 & 30.78 \\
+ TTSV & 74.40 & 22.80 & 29.80 & 65.00 & 48.00 \\
$\Delta$(EM)    & +15.00 & +7.80 & +16.00 & +17.50 & +14.08 \\
$\Delta$(TTSV)     & +23.40 $^{\uparrow\text{45.88\%}}$ & +9.90$^{\uparrow\text{76.74\%}}$ & +13.10$^{\uparrow\text{78.44\%}}$ & +22.50$^{\uparrow\text{52.94\%}}$ & +17.22$^{\uparrow\text{55.95\%}}$ \\
\midrule
LLaMA-3.1-8B      & 50.60 & 17.30 & 11.60 & 25.00 & 26.13 \\
+ TTSV & 49.40 & 17.60 & 14.20 & 25.00 & 26.55 \\
$\Delta$(EM)    & +0.20 & +1.90 & -1.60 & +0.30 & +0.20 \\
$\Delta$(TTSV)     & -1.20$^{\downarrow\text{2.37\%}}$ & +0.30$^{\uparrow\text{1.73\%}}$ & +2.60$^{\uparrow\text{22.41\%}}$ & +0.00$^{\uparrow\text{0.00\%}}$ & +0.42$^{\uparrow\text{1.61\%}}$ \\
\bottomrule
\end{tabular}
\caption{Performance Comparison of TTSV against EM on the Qwen2.5-Math-7B and LLaMA-3.1-8B Models.}
\label{EM}
\end{table*}

\section{Method}
Herein, we detail our proposed Model Whisper, which is designed to guide the LLM by optimizing a set of external, lightweight vectors, TTSV, on unlabeled test data, enhancing LLM's performance on specific tasks without modifying any of the model's internal parameters. Figure \ref{framework} presents an overview of the entire framework.

\subsection{Formulation}
Given a pre-trained LLM, denoted as $M(\cdot; \theta)$, with its parameters $\theta$ entirely frozen, our objective is to enhance its performance on an unlabeled test dataset $\mathcal{D}_{\text{test}} = \{X_1, X_2, \dots, X_N\}$.

To this end, we introduce the TTSV, learnable continuous vectors denoted as $\boldsymbol{V}_{\text{steer}} \in \mathbb{R}^{L \times d}$. Here, $L$ represents the sequence length of the steering vectors (a small hyperparameter, e.g., $L=20$), and $d$ is the model's embedding dimension. The operational mechanism of TTSV is straightforward: for any input text $X$, we first retrieve its embedding sequence, $E(X)$. The optimized tensor, $\boldsymbol{V}_{\text{steer}}$, is then prepended to this sequence, creating an augmented input $E' = [\boldsymbol{V}_{\text{steer}}; E(X)]$. This augmented sequence is subsequently fed into the frozen LLM $M$ to generate the final output. In this manner, $\boldsymbol{V}_{\text{steer}}$ modulates the model's initial computational state, thereby steering its subsequent reasoning process without altering its intrinsic parameters.

\subsection{Optimization}

Our optimization strategy is based on a core premise: a model proficient in a specific task should exhibit high certainty in its predictions, which corresponds to low output entropy. By inverting this relationship, we adopt entropy minimization as our optimization objective. We aim to actively steer the model towards this state of high certainty by reducing its output entropy, thereby activating its inherent, task-relevant reasoning abilities. 

When the model processes the augmented input embedding $E' = [\boldsymbol{V}_{\text{steer}}; E(X)]$, it autoregressively generates a token sequence $Y = (y_1, y_2, \dots, y_T)$. For the $t$-th step of this generation, the conditional entropy $H_t$ is defined as:
\begin{equation}
H_t = - \sum_{v \in \mathcal{V}} p(v | y_{<t}, E') \log p(v | y_{<t}, E').
\end{equation}
To focus the optimization on the model's generative capabilities, our final loss function is defined as the average token-level entropy across all non-padding generated tokens within a mini-batch. For a mini-batch of $B$ samples, the loss function $\mathcal{L}$ is computed as:
\begin{equation}
\mathcal{L}(\boldsymbol{V}_{\text{steer}}) = \frac{\sum_{i=1}^{B} \sum_{t \in \mathcal{I}_i} H_{i,t}}{\sum_{i=1}^{B} |\mathcal{I}_i|},
\label{eq:final_loss}
\end{equation}
where $\mathcal{I}_i$ denotes the set of all non-padding generated token positions for the $i$-th sample, and $H_{i,t}$ is the conditional entropy for sample $i$ at time step $t$.

This loss function is differentiable with respect to $\boldsymbol{V}_{\text{steer}}$. With backpropagation, we compute and update gradients only for $\boldsymbol{V}_{\text{steer}}$, while the model parameters $\theta$ remain frozen. This unsupervised entropy minimization process, without relying on any external labels, effectively compels the model to converge to a more deterministic internal configuration, thereby unlocking its potential for the task at hand.

\subsection{Inference}

Once TTSV optimized via the process described above, we obtain fixed steering vectors, denoted as $\boldsymbol{V}_{\text{steer}}^*$, which is tailored to the current test distribution. During the inference phase, these optimized steering vectors are used for all samples in the test set.

For any given instance $X_i \in \mathcal{D}_{\text{test}}$, the inference process is straightforward: the fixed steering tensor $\boldsymbol{V}_{\text{steer}}^*$ is prepended to its corresponding embedding sequence $E(X_i)$ to construct the augmented input $E'_i = [\boldsymbol{V}_{\text{steer}}^*, E(X_i)]$. This augmented sequence is then fed into the frozen LLM to generate the final prediction.

This ``optimize-once, apply-to-all'' paradigm renders TTSV a true plug-and-play component. The overhead at inference time is negligible, limited only to the processing of $L$ additional vectors at the input layer. Consequently, it enhances model performance with virtually no impact on inference speed.

\section{Experiment}

\subsection{Experimental Setup}

\subsubsection{Models.}

To demonstrate the broad applicability of our method, we conducted experiments on both base models and reasoning-enhanced models. For base models, we selected models from both the Qwen and LLaMA series to ensure architectural diversity. Specifically, we used Qwen2.5-Math \cite{yang2024qwen2} in two different scales (1.5B and 7B) and the LLaMA-3.1-8B-Instruct model \cite{grattafiori2024llama}. For the reasoning-enhanced category, we chose the latest Qwen3-4B model \cite{yang2025qwen3} operating in its `thinking mode' to evaluate our method's effectiveness on state-of-the-art architectures.

\subsubsection{Benchmarks.}

Our evaluation is conducted on a diverse suite of benchmarks to test mathematical and general reasoning.
For mathematical reasoning, we employ widely-used datasets such as MATH500 \cite{hendrycks2021measuring_math500}, Minerva Math \cite{lewkowycz2022solving_minervamath}, Olympiad Bench \cite{he2024olympiadbench}, and AMC23. To assess performance on problems of higher difficulty, our evaluation also includes AIME24 \cite{li2024numinamath}, derived from the American Invitational Mathematics Examination 2024, which tests advanced, competition-style mathematical problem-solving.
In addition, to assess the generalization of our method beyond the mathematical domain, we utilize GPQA Diamond \cite{rein2024gpqa}, a benchmark comprising graduate-level scientific questions in fields like physics and biology, specifically designed to be resistant to web-search retrieval.

\subsubsection{Baselines.}
We benchmark our method against two representative test-time adaptation methods: EM (Entropy Minimization) \cite{gao2025em} and SLOT \cite{hu2025slot}. EM represents a full fine-tuning approach that adjusts all model parameters at test time, while SLOT is a parameter-efficient method that performs sample-specific fine-tuning on the output vector of the model's last hidden layer. We validate the effectiveness of our method through comparison with these approaches.

\subsubsection{Implementation Details.}

We set the length $L$ of TTSV to 20. The TTSV was optimized for 20 epochs using the AdamW optimizer. We employed a linear learning rate schedule, with the rate initialized at 1e-3 and decaying to 1e-5. Other optimization hyperparameters included a weight decay of 1e-8, an epsilon of 1e-5, and a training batch size of 16. During inference, we used a batch size of 64, and the maximum generation length was set to 3072 tokens. All experiments were conducted on NVIDIA A800 GPUs. Further implementation details are provided in Section 7.

\subsection{Main Results}

\begin{table*}[ht]
\centering
\begin{tabular}{lcccc}
\toprule
\textbf{Model} & \textbf{AIME24} & \textbf{MATH500} & \textbf{GPQA} & \textbf{Avg.} \\
\midrule
Qwen2.5-Math-1.5B    & 6.67  & 39.40 & 27.27 & 24.45  \\
+ TTSV & 6.67 & 68.40 & 29.80 & 34.96  \\
$\Delta$(SLOT)    & +3.33 & +0.00 & -0.51 & +0.94  \\
$\Delta$(TTSV)     & +0.00$^{\uparrow\text{0.00\%}}$ & +29.00$^{\uparrow\text{73.60\%}}$ & +2.53$^{\uparrow\text{9.28\%}}$ & +10.51$^{\uparrow\text{42.99\%}}$  \\

\midrule
Qwen2.5-Math-7B      & 13.33 & 51.00 & 28.79 & 31.04  \\
+ TTSV & 23.33 & 74.40 & 31.82  & 43.18  \\
$\Delta$(SLOT)    & +6.67 & +1.20 & +7.07 & +4.98  \\
$\Delta$(TTSV)     & +10.00$^{\uparrow\text{75.02\%}}$ & +23.40$^{\uparrow\text{45.88\%}}$ & +3.03$^{\uparrow\text{10.52\%}}$  & +12.14$^{\uparrow\text{39.12\%}}$  \\

\bottomrule
\end{tabular}
\caption{Performance Comparison of TTSV against SLOT on the Qwen2.5-Math-1.5B and Qwen2.5-Math-7B Models.}
\label{SLOT}
\end{table*}

\begin{table*}[ht]
\centering
\begin{tabular}{lcccccccc}
\toprule
\textbf{Model} & \textbf{MATH500} & \textbf{AMC23} & \textbf{Minerva Math} & \textbf{Olympiad} & \textbf{AIME24}  & \textbf{GPQA} & \textbf{Avg.} \\
\midrule
Qwen3-4B      & 51.80 & 37.50 & 24.60 & 18.10 & 56.67 & 41.92 & 38.43  \\
+ TTSV & 60.20 & 40.00 & 25.40  & 30.70 & 60.00 & 47.98 & 44.05  \\
$\Delta$     & +8.40$^{\uparrow\text{16.22\%}}$ & +2.50$^{\uparrow\text{6.67\%}}$ & +0.80$^{\uparrow\text{3.25\%}}$  & +12.60$^{\uparrow\text{69.61\%}}$ & +3.33$^{\uparrow\text{5.88\%}}$ & +6.06$^{\uparrow\text{14.46\%}}$ & +5.62$^{\uparrow\text{14.62\%}}$  \\
\bottomrule
\end{tabular}
\caption{Performances of TTSV on Reasoning Model.}
\label{reasoning}
\end{table*}

\subsubsection{Comparison with Full Parameter TTA (EM).}
Table \ref{EM} shows the experimental results of TTSV and the full fine-tuning EM on the Qwen2.5-Math-7B model and LLaMA-3.1-8B model. On the Qwen2.5-Math-7B model, TTSV demonstrates significant efficacy, achieving an average absolute improvement of +17.22\%, which corresponds to a 55.95\% relative gain. This result surpasses the +14.08\% average gain from EM. On the LLaMA-3.1-8B model, while the gains from both methods were modest, TTSV still achieved a slightly higher average improvement than EM. These results suggest that our approach provides an effective path to enhance model performance while greatly reducing the computational overhead typically associated with full fine-tuning during test-time adaptation.

\subsubsection{Comparison with Parameter-Efficient TTA (SLOT).}
We further analyze our method by comparing it with SLOT, a parameter-efficient TTA approach. While both methods keep core model parameters frozen and operate on a small number of tunable parameters, they differ in their point of intervention. SLOT performs a ``post-computation adjustment'' by modifying the output features of the model's final layer, acting akin to an output filter. In contrast, our TTSV implements a ``pre-computation steering'' mechanism, guiding the model's entire reasoning process from the outset by prepending optimized vectors to the input.

As shown in Table \ref{SLOT}, TTSV achieves superior performance on both the Qwen2.5-Math-1.5B and 7B models.

However, the two methods employ different adaptation strategies. SLOT's ``sample-specific'' approach offers flexibility, as it can adapt without requiring the entire test set, albeit at the cost of inference latency due to per-sample optimization. We acknowledge the value of this strategy in certain scenarios. To provide a direct comparison, we also conducted a sample-specific experiment with TTSV on the Qwen-2.5-math-1.5B model. As shown in Table \ref{sampel_specific}, the results demonstrate that even in this setting, TTSV still outperforms SLOT. The superior outcomes under both adaptation strategies suggest that guiding the model's internal state early in the computational graph may be a more effective way to activate latent abilities than adjusting its final output.

Therefore, based on a comprehensive consideration of performance, stability, and inference efficiency, we adopt the ``optimize-once, apply-to-all'' paradigm as the primary approach in this work.

\begin{table}[ht]
\centering
\begin{tabular}{lccc}
\toprule
\textbf{Model} & \textbf{AIME24} & \textbf{MATH500} & \textbf{GPQA}  \\
\midrule
Baseline    & 6.67  & 39.40 & 27.27   \\
+ TTSV & 10.00 & 43.40 & 27.78   \\
$\Delta$(SLOT)    & +3.33 & +0.00 & -0.51   \\
$\Delta$(TTSV)     & +3.33 & +4.00 & +0.51   \\

\bottomrule
\end{tabular}
\caption{Performance Comparison of TTSV (sample-specific) against SLOT on the Qwen2.5-Math-1.5B Model.}
\label{sampel_specific}
\end{table}

\subsubsection{TTSV for Reasoning Model.}

To assess the applicability of our method to state-of-the-art LLMs, we further evaluated TTSV on a reasoning-enhanced model. Such models, having been specifically post-trained for structured, complex reasoning, present a new challenge for our approach. The task for TTSV is no longer just to activate a latent ability, but to precisely steer an already sophisticated reasoning mechanism without disrupting its inherent strengths.
We selected the latest Qwen3-4B model (thinking mode) for this purpose and compared its performance with and without TTSV. As detailed in Table \ref{reasoning}, TTSV demonstrates performance gains across all benchmarks. The effect is particularly notable on Olympiad Bench, where it achieves a substantial relative increase of 69.61\%. Across all datasets, TTSV achieves an average relative improvement of 14.62\%.
These results indicate that TTSV, despite a lightweight external component, can effectively synergize with advanced reasoning models, helping to further focus and refine their reasoning pathways. This confirms that our approach is not only effective for base models but also holds significant relevance and potential for the frontier of LLM development.

\subsection{Ablation Studies}

\subsubsection{Analysis of Training Dynamics.}
\begin{figure}[ht]
\centering
\includegraphics[width=0.95\columnwidth]{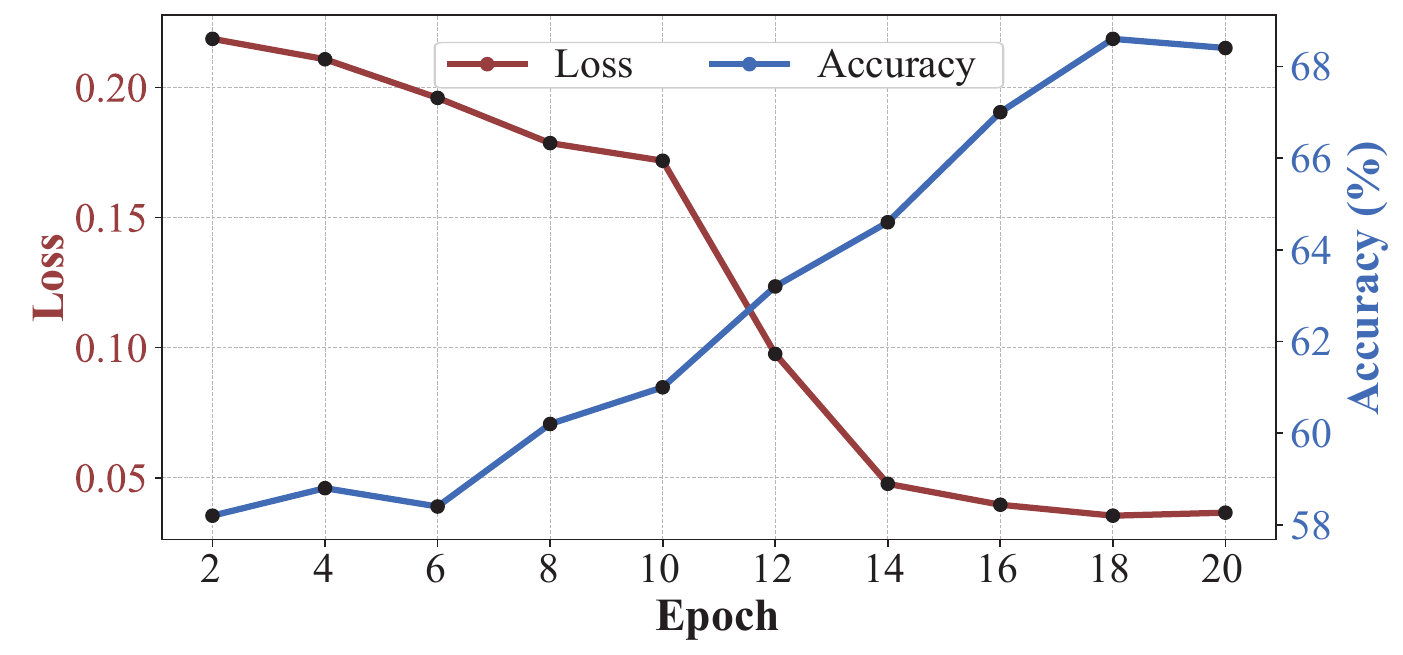} 
\caption{Training dynamics of Qwen2.5-Math-1.5B on MATH500. The plot illustrates the change in training loss (left y-axis, red line) and accuracy (right y-axis, blue line) over 20 epochs.}
\label{loss-acc}
\end{figure}

To investigate the training dynamics of TTSV, we tracked the relationship between the entropy-based loss and task accuracy. As depicted in Figure \ref{loss-acc} , as the training loss steadily decreases, the model's accuracy consistently rises until both metrics converge to a plateau. Notably, this process is stable and does not exhibit the performance degradation at later training stages observed in EM.

We attribute this stability to our approach's fundamental design, which operates on an external component while leaving the model's internal parameters untouched. In contrast, methods like EM, which directly modify the core weights of the model, are susceptible to overfitting on the test distribution. The entropy minimization objective can push the model into a state of over-confidence, leading to overly deterministic and shallow reasoning \cite{zhang2025no}, where the model learns to be certain but incorrect.

By keeping the LLM's parameters frozen, TTSV preserves the integrity of the model's core reasoning abilities and its vast knowledge base. The optimization process is thereby constrained to a search for an optimal activation signal within the model's existing latent space, rather than a reshaping of the space itself. This inherently guards against overfitting and the risk of catastrophic forgetting, ensuring the model's foundational capabilities remain intact.

\subsubsection{Impact of TTSV Length}

\begin{figure}[ht]
\centering
\includegraphics[width=0.95\columnwidth]{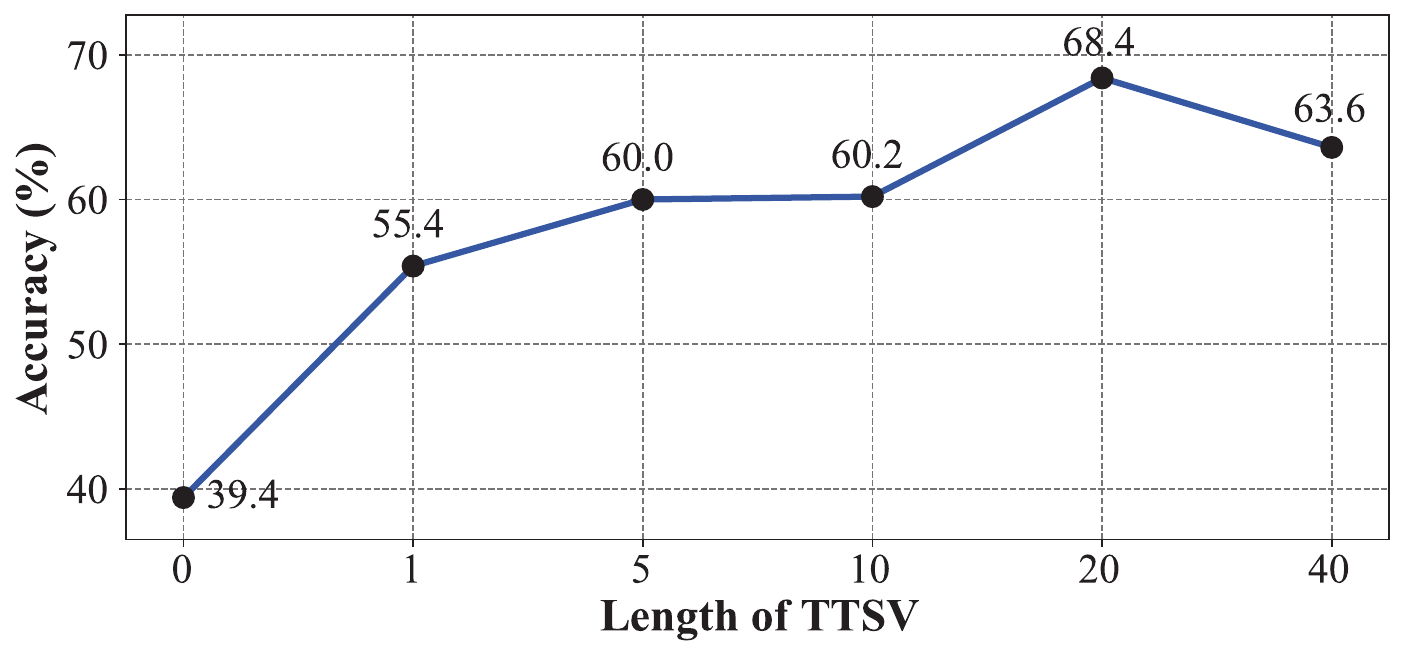} 
\caption{The effect of TTSV length on the accuracy of Qwen2.5-Math-1.5B on the MATH500 benchmark.}
\label{length-acc}
\end{figure}

The length of the TTSV, L , is a key hyperparameter that determines the representational capacity of the steering signal. We evaluated its impact by varying the length, setting $L\in \{1,5,10,20,40\}$. The results, presented in Figure \ref{length-acc} , yield two important findings.

First, the relationship between TTSV length and performance is non-monotonic. Accuracy improves as L increases from 1 to 20, with $L=20$ yielding the best results, but performance declines at $L=40$. This suggests a trade-off: a short prefix may have insufficient capacity to encode an effective steering signal, while an excessively long prefix increases the number of parameters to be optimized and may introduce noise or unfavorably alter the attention dynamics over the primary task input. Thus, an optimal length exists that balances expressive capacity with optimization stability.

Second, it is noteworthy that a TTSV of length $L=1$ still provides a substantial improvement over the baseline. This finding powerfully demonstrates the high information density of the model's continuous embedding space. It suggests that a single, well-optimized vector is sufficient to convey a potent steering command to the model. The ability to guide a model's behavior with such a minimal and continuous signal hints at a promising area for further investigation into efficient model control.

\subsubsection{Analysis of Cross-Distribution Generalization}

\begin{figure*}[ht]
\centering
\includegraphics[width=1.0\textwidth]{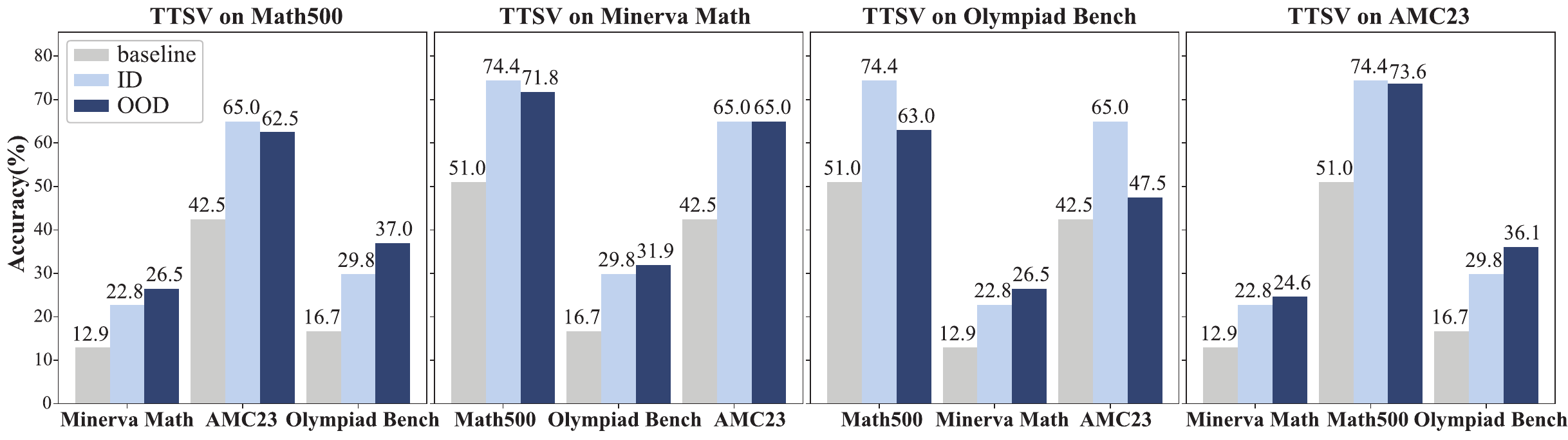} 
\caption{Cross-distribution generalization of TTSV. The figure compares the performance of the Qwen2.5-Math-7B model with In-Distribution (ID) and Out-of-Distribution (OOD) TTSV applications across four math datasets.}
\label{cross_domain}
\end{figure*}

To assess the transferability of the steering signals learned by TTSV, we conducted a cross-distribution generalization experiment, with results presented in Figure \ref{cross_domain}. A consistent finding from this experiment is that applying an Out-of-Distribution (OOD) TTSV, one trained on a different distribution, always yields substantial performance gains over the baseline. For instance, the TTSV trained on Math500 improved accuracy on AMC23 from 42.5\% to 62.5\%. This indicates that the learned steering signal captures a significant degree of cross-distribution utility, rather than simply overfitting to the source distribution.

Building on this, we observed an interesting phenomenon where, on some tasks, the OOD application outperformed the In-Distribution (ID) TTSV optimized for the target. A clear example is seen when the TTSV trained on Math500 is applied to the Minerva Math task, recording an accuracy of 26.5\%, which surpasses the 22.8\% from the ID-TTSV trained directly on Minerva Math. We attribute this to the possibility that different training distributions shape TTSVs to specialize in distinct reasoning capabilities. While an ID-TTSV must find a comprehensive solution for all problems within its distribution, an OOD-TTSV may happen to specialize in activating a core ability that is particularly critical for the target task, thus yielding superior performance.

\section{Theoretical Analysis}

We theoretically elaborate on the core mechanism of our proposed TTSV method. Our central thesis is that by prepending a learnable vector to the model's input, TTSV introduces an initial bias into the computation. This signal is then propagated and amplified through the deep network, systematically steering the model's computational trajectory towards a desired target state.

We begin by analyzing the precise effect of TTSV within the first attention layer of the model. As derived in Section 8, ignoring position encoding, when a learnable vector $\boldsymbol{p}$ is introduced, the new output $\boldsymbol{t}'_i$ at any content position $i$ can be expressed exactly as a linear combination of the original output $\boldsymbol{t}_i$ and a bias direction $\boldsymbol{b}$:
\begin{equation}
    \boldsymbol{t}'_i = (1 - \alpha_i) \boldsymbol{t}_i + \alpha_i \boldsymbol{b}.
    \label{eq:bias_effect}
\end{equation}
The bias direction $\boldsymbol{b} = \boldsymbol{W}_V \boldsymbol{p}$ is determined solely by the learnable vector $\boldsymbol{p}$, while the bias strength $\alpha_i$ is the attention weight from position $i$ to $\boldsymbol{p}$. Eq. ~\ref{eq:bias_effect} demonstrates that TTSV introduces a linear guiding signal in the first layer, shifting the model's activation state from its original trajectory.

This initial guiding signal propagates through the model's deep network. The computation at each layer operates on the already-guided trajectory from the previous one, which allows a small initial offset to be significantly amplified after passing through multiple nonlinear transformations. Our experiments (in Section 8) also corroborate this trend of ``bias amplification'', empirically showing that the magnitude of the activation shift induced by TTSV increases with network depth.

\begin{figure}[ht]
\centering
\includegraphics[width=1.0\columnwidth]{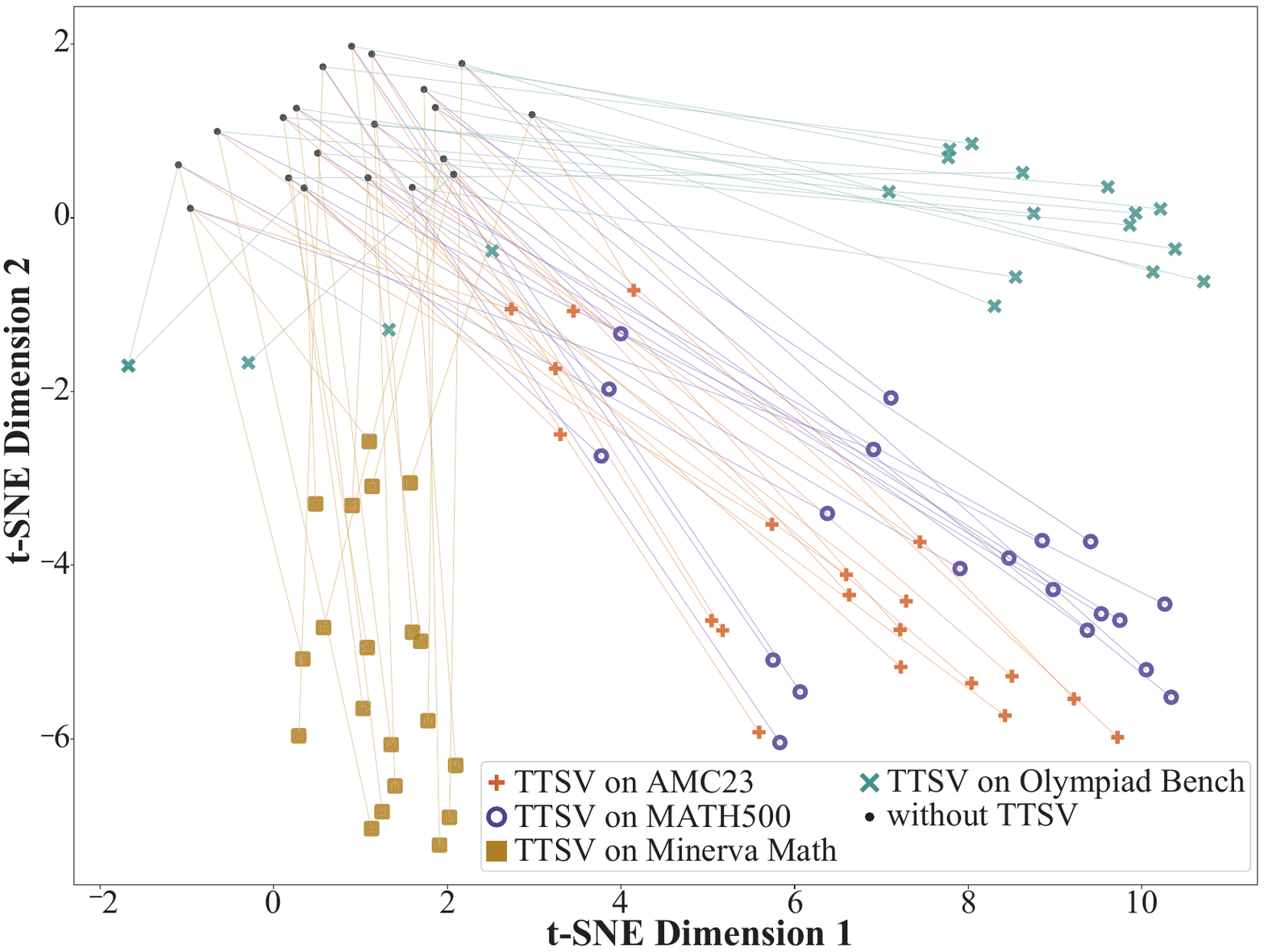} 
\caption{t-SNE visualization of the final layer activations from the Qwen2.5-math-7B model on 20 problems sampled from Minerva Math.}
\label{attention}
\end{figure}

The cumulative effect of this process is manifested in the final layer's activations, as visualized by the t-SNE plot in Figure~\ref{attention}. In the figure, the baseline activations (black dots) are systematically pushed towards new, task-specific regions (colored markers) under the influence of TTSV. The vectors connecting the baseline and task-specific points vividly illustrate this aggregate guiding force. This provides strong evidence that TTSV, by injecting a controllable bias early in the computational flow and leveraging the model's depth to amplify this signal, successfully steers the model's general-purpose capabilities into the specialized states required for specific tasks. Furthermore, we observe an overlap in the state distributions guided by TTSV vectors trained on different tasks. This observation may explain the strong OOD performance of TTSV reported in Section 4.3, as it suggests the model is guided towards a shared, robust reasoning subspace rather than narrow, task-specific solutions.

\section{Conclusion}

In this paper, we introduced a novel method to enhance the reasoning capabilities of Large Language Models without modifying their internal parameters. By optimizing a small set of input vectors to minimize output entropy, TTSV effectively steers the model towards a state of higher confidence. Our comprehensive experiments and theoretical analyses demonstrate that this lightweight, plug-and-play method is capable of guiding the model's trajectory towards a target state from the very outset, achieving significant performance gains on both base and reasoning-enhanced models, and exhibiting remarkable generalization across different tasks. 

This work illuminates a promising and efficient path, as an alternative to full fine-tuning, to unlock the latent potential of LLMs. Furthermore, this work suggests several potential open questions. First, beyond entropy minimization, the exploration of alternative or combined objective functions remains an important open question. Second, integrating external guidance mechanisms like TTSV with training methods such as reinforcement learning might offer new perspectives on model optimization. Finally, extending the application of this approach from LLMs to Multimodal Large Language Models (MLLMs) also warrants investigation. We hope this work will inspire further research into resource-efficient ways to guide and improve existing models.

\bibliography{aaai2026}

\clearpage 
\section{Implementation Details}
\label{Implementation Details}
This section provides a comprehensive overview of the experimental settings used in our study, including the initialization of TTSV, the hyperparameters for both the training and evaluation phases, and the learning rate configurations for different models.

\subsection{TTSV Initialization}
The initialization strategy for the TTSV was adapted based on the model's sensitivity to initial parameter states.

\begin{itemize}
    \item For Qwen Series Models: We employed a standard random initialization for TTSV. Specifically, the parameters were drawn from a standard normal distribution, $\mathcal{N}(0, 1)$. This approach proved to be effective and stable for the Qwen architecture.

    \item For Llama Series Models: We observed that Llama models are more sensitive to the initialization of trainable parameters. A purely random initialization could lead to suboptimal convergence states. To mitigate this, we adopted a data-driven initialization method. This was achieved by processing the entire training dataset to compute the mean and variance of the corresponding token embeddings. These empirical statistics were then used as the parameters ($\mu$ and $\sigma^2$) for the normal distribution from which the initial TTSV weights were sampled. This strategy provides a more informed starting point, aligning the initial state of TTSV with the embedding space of the target data.
\end{itemize}

\subsection{Generation Hyperparameters}
We configured distinct sets of generation hyperparameters for the training and evaluation phases. The core parameters for each phase are summarized in Table~\ref{tab:gen_params}. For the evaluation phase, we switch to greedy decoding to ensure deterministic and reproducible outputs.

\begin{table}[ht]
\centering
\caption{Core generation hyperparameters for the training and evaluation phases.}
\label{tab:gen_params}
\resizebox{\columnwidth}{!}{%
\begin{tabular}{lcc}
\toprule
\textbf{Parameter} & \textbf{Training Phase} & \textbf{Evaluation Phase} \\
\midrule
\textbf{do sample} & True & False \\
\textbf{top p} & 0.95 & - \\
\textbf{temperature} & 0.5 & - \\
\textbf{repetition penalty} & 1.15 & - \\
\bottomrule
\end{tabular}
}
\end{table}

The \textbf{max new tokens} hyperparameter was dynamically adjusted to accommodate varying output lengths. During the training phase, it was set to \textbf{256} for the Qwen2.5 series and Llama models, increased to \textbf{1024} for the Qwen3-4B model to account for its ``thinking'' process, and further extended to \textbf{2048} for the particularly challenging AIME24 dataset. In the evaluation phase, this value was generally set to \textbf{3072}. For the AIME24 dataset evaluation with the Qwen3 model, it was significantly increased to \textbf{20480}, following the set in \cite{zhang2025no} .

\subsection{Learning Rate Configuration}
The learning rate was carefully selected for each model family to ensure stable and effective training.

\begin{itemize}
    \item For Qwen Series Models: A learning rate of $1\times10^{-3}$ was used, which resulted in robust convergence.
    
    \item For Llama Series Models: Due to the data-driven initialization of TTSV, we found that a high learning rate like $1\times10^{-3}$ caused the entropy of the output distribution to collapse prematurely within the first few training steps, hindering effective learning. To address this, we employed a much smaller learning rate of $5\times10^{-6}$. This conservative rate facilitated a more gradual and stable optimization process, allowing the model to effectively fine-tune the TTSV parameters without destabilizing the generation process.
\end{itemize}

\begin{figure*}[ht!]
\centering
\includegraphics[width=0.8\textwidth]{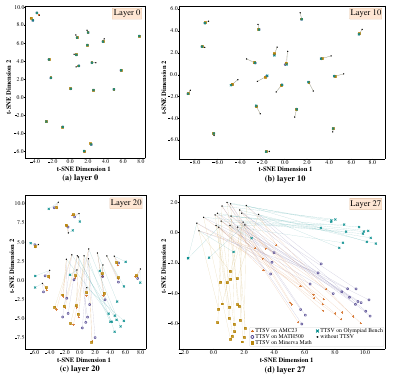}
\caption{The magnitude of activation shift induced by TTSV in different layers, (a) layer 0, (b) layer 10, (c) layer 20, (d) layer 27. The distributions show that the shift magnitude increases with network depth.}
\label{fig:bias_amplification}
\end{figure*}

\section{Theoretical Analysis of TTSV}

In this section, we provide a detailed theoretical analysis of TTSV. We first formally derive the mechanism by which TTSV introduces a bias in the first attention layer. We then present empirical evidence demonstrating how this initial bias is amplified through the network's depth.

\subsection{Derivation of the Linear Bias in the First Attention Layer}
We analyze the effect of prepending a single learnable vector $\boldsymbol{p}$ to a content sequence of tokens $\boldsymbol{x}_1, \dots, \boldsymbol{x}_L$. Let the combined input sequence to the first attention layer be $(\boldsymbol{p}, \boldsymbol{x}_1, \dots, \boldsymbol{x}_L)$.

The output of a standard attention head at a content position $i$ (where $i \in \{1, \dots, L\}$) is the weighted sum of value vectors:
\begin{equation}
    \boldsymbol{t}_i = \sum_{j=1}^{L} A_{ij} \boldsymbol{W}_V \boldsymbol{x}_j,
\end{equation}
where $A_{ij}$ is the attention weight from position $i$ to $j$, and $\boldsymbol{W}_V$ is the value matrix. The attention weights are computed via a softmax over the attention scores:
\begin{equation}
    A_{ij} = \frac{\exp(s_{ij})}{\sum_{k=1}^{L} \exp(s_{ik})},
\end{equation}
where $s_{ik} = (\boldsymbol{W}_Q \boldsymbol{x}_i)^\top (\boldsymbol{W}_K \boldsymbol{x}_k) / \sqrt{d_k}$ is the attention score.

When the learnable vector $\boldsymbol{p}$ is introduced at position 0, the output at content position $i$, denoted as $\boldsymbol{t}'_i$, becomes:
\begin{equation}
    \boldsymbol{t}'_i = A'_{i0} \boldsymbol{W}_V \boldsymbol{p} + \sum_{j=1}^{L} A'_{ij} \boldsymbol{W}_V \boldsymbol{x}_j.
\end{equation}
Here, $A'_{ij}$ are the new attention weights, now computed over the extended sequence. The presence of $\boldsymbol{p}$ only adds a new term $s_{i0}$ to the denominator of the softmax for all content-to-content attention weights $A'_{ij}$ ($j \ge 1$). This means the relative attention between content tokens remains unchanged, but their total weight is scaled down. Let $\alpha_i = A'_{i0}$ be the attention from position $i$ to the learnable vector $\boldsymbol{p}$. The new attention weight $A'_{ij}$ can be expressed in terms of the original weight $A_{ij}$ and $\alpha_i$:
\begin{equation}
    A'_{ij} = (1 - \alpha_i) A_{ij} \quad \text{for } j \in \{1, \dots, L\}.
\end{equation}
This holds because $\sum_{j=1}^{L} A'_{ij} = 1 - A'_{i0} = 1 - \alpha_i$, and the relative proportions within the sum are preserved.

Substituting this back into the equation for $\boldsymbol{t}'_i$:
\begin{align}
    \boldsymbol{t}'_i &= \alpha_i \boldsymbol{W}_V \boldsymbol{p} + \sum_{j=1}^{L} (1 - \alpha_i) A_{ij} \boldsymbol{W}_V \boldsymbol{x}_j \\
    &= \alpha_i \boldsymbol{W}_V \boldsymbol{p} + (1 - \alpha_i) \sum_{j=1}^{L} A_{ij} \boldsymbol{W}_V \boldsymbol{x}_j \\
    &= \alpha_i \boldsymbol{W}_V \boldsymbol{p} + (1 - \alpha_i) \boldsymbol{t}_i.
\end{align}
By defining the bias direction as $\boldsymbol{b} = \boldsymbol{W}_V \boldsymbol{p}$, and noting that $\alpha_i$ is the bias strength, we arrive at the expression presented in the main text:
\begin{equation}
    \boldsymbol{t}'_i = (1 - \alpha_i) \boldsymbol{t}_i + \alpha_i \boldsymbol{b}.
    \label{eq:appendix_bias_effect}
\end{equation}
This derivation formally shows that in the first layer, TTSV's effect is a linear interpolation between the original activation $\boldsymbol{t}_i$ and a constant bias direction $\boldsymbol{b}$, with the interpolation factor $\alpha_i$ determined by the attention paid to the learnable vector.

\begin{figure*}[ht!]
\centering
\includegraphics[width=0.95\textwidth]{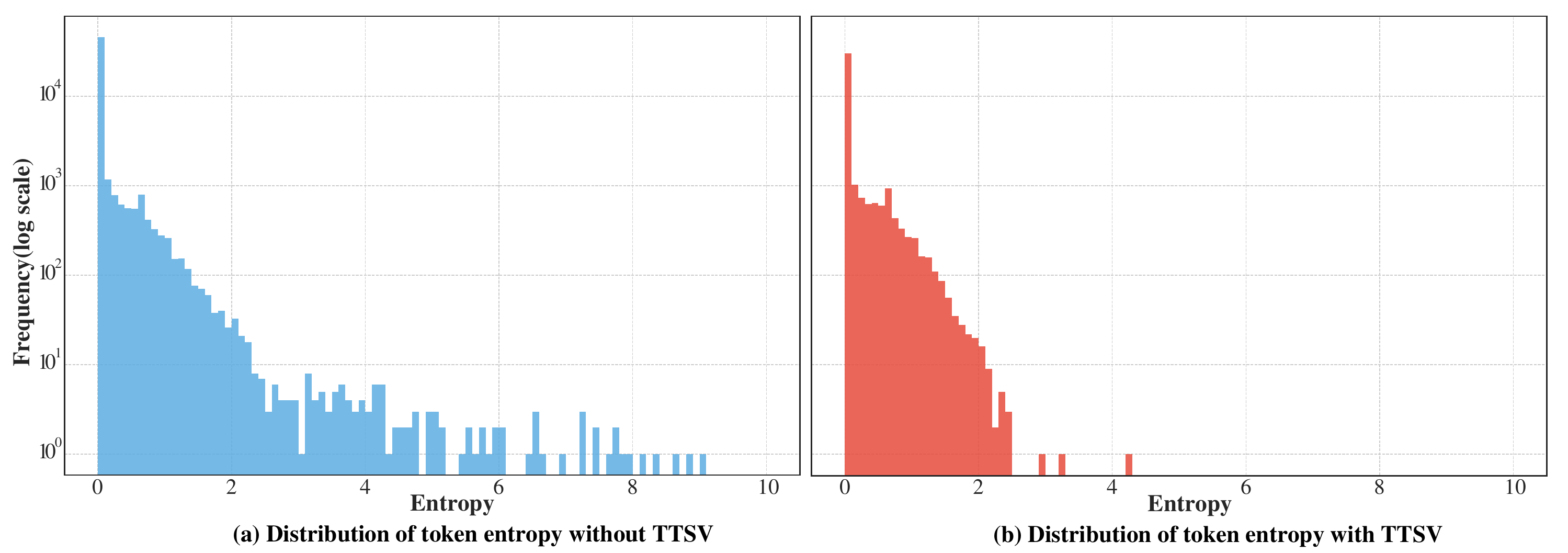}
\caption{Distribution of token entropy before (a) and after (b) applying TTSV. The analysis was performed on the \textbf{Qwen2.5-math-7B} model using outputs from the AMC23 dataset.}
\label{fig:entropy_dist}
\end{figure*}

\subsection{Empirical Validation of Bias Amplification}
To visualize the bias amplification effect, we analyzed the output activations of specific attention layers within the Qwen2.5-math-7B model. The analysis was conducted on 20 problems randomly sampled from the Minerva Math dataset. Figure~\ref{fig:bias_amplification} presents the t-SNE visualization of these activations at Layers 0, 10, 20, and 27. The visualization clearly shows that the small initial shift between the baseline (without TTSV) and TTSV-guided activations at Layer 0 is progressively magnified through subsequent attention layers. By Layer 27, this results in a significant divergence, empirically confirming that the initial bias is amplified with network depth.

\section{Effect of TTSV on Output Entropy}

To further understand the mechanism of TTSV, we analyzed its effect on the entropy of the model's output token distribution. Figure~\ref{fig:entropy_dist} illustrates a histogram comparing the token entropy distribution with and without the application of TTSV.

The baseline model (without TTSV) exhibits a broad entropy distribution with a long tail, indicating a significant presence of high-entropy tokens. This suggests that the model often faces high uncertainty during the generation process. In contrast, the application of TTSV fundamentally reshapes this distribution. The high-entropy tail is effectively suppressed, and the distribution becomes highly concentrated in the low-entropy region. This indicates that TTSV guides the model towards a more deterministic and confident generation path, significantly reducing uncertainty at each step.

\end{document}